\documentclass{article}

\usepackage{microtype}
\usepackage{graphicx}
\usepackage{subfigure}
\usepackage{booktabs} 

\usepackage{hyperref}

\usepackage[accepted]{icml2025}
\setcitestyle{numbers}

\usepackage{amsmath}
\usepackage{amssymb}
\usepackage{mathtools}
\usepackage{amsthm}

\usepackage[capitalize,noabbrev]{cleveref}

\theoremstyle{plain}

\theoremstyle{definition}

\theoremstyle{remark}

\usepackage[textsize=tiny]{todonotes}

\newcommand{\dataset}{BiasLab}
\usepackage[most]{tcolorbox}

\icmltitlerunning{Submission and Formatting Instructions for ICML 2025}
\icmltitlerunning{\dataset: Explainable Political Bias Detection with Human Annotations and Rationale}

\begin{document}

\twocolumn[
\icmltitle{\dataset: Toward Explainable Political Bias Detection with Dual-Axis Human Annotations and Rationale Indicators}



\icmlsetsymbol{equal}{*}

\begin{icmlauthorlist}
\icmlauthor{KMA Solaiman}{yyy} 
\end{icmlauthorlist}

\icmlaffiliation{yyy}{Department of CSEE, University of Maryland, Baltimore County, MD, USA}

\icmlcorrespondingauthor{KMA Solaiman\\}{ksolaima@umbc.edu}

\icmlkeywords{political bias detection, explainable AI, interpretable NLP, dataset annotation, dual-axis sentiment, rationale annotations, human annotation, large language models, perception, crowdsourcing}

\vskip 0.3in
]



\printAffiliationsAndNotice{}  

\begin{abstract}
We present \textbf{\dataset}, a dataset of 300 political news articles annotated for perceived ideological bias. These articles were selected from a curated 900-document pool covering diverse political events and source biases.  Each article is labeled by crowdworkers along two independent scales, assessing sentiment toward the Democratic and Republican parties, and enriched with rationale indicators adapted from media literacy guidelines. The annotation pipeline incorporates targeted worker qualification and was refined through pilot-phase analysis. We quantify inter-annotator agreement, analyze misalignment with source-level outlet bias, and organize the resulting labels into interpretable subsets. We complement human annotations with schema-constrained large language model (LLM) labeling to compare and align human and machine interpretability. 
These annotations captured mirrored asymmetries, especially in misclassifying subtly right-leaning content.
We define two modeling tasks: \textit{perception drift} prediction and \textit{rationale type} classification, and report baseline performance to illustrate the challenge of explainable bias detection. Our analysis reveals notable disagreement patterns and perception drift, underscoring the subjectivity inherent in bias perception. We release the dataset, annotation schema, and baseline code, enabling feedback-driven alignment research using structured human labels and rationales for political bias perception.

Code \& data: \textit{\url{https://github.com/ksolaiman/PoliticalBiasCorpus}}
\end{abstract}

\section{Introduction}
Political news reporting often reflects ideological leanings not just through overt opinion, but also through subtle cues such as selective emphasis, framing, or omission. While such framing is not inherently deceptive, it can reinforce partisan divides, obscure nuance, and influence public understanding of policy issues \cite{entman2007framing, lazer2018fakenews}. 
Detecting such bias is not only a classification problem but also an explainability challenge: practitioners need models that can surface why a story reads as left- or right-leaning, not just that it does.
While computational approaches to bias detection have made notable progress, most rely on coarse-grained supervision such as outlet-level bias labels \cite{allsides, adfontes} or focus narrowly on framing in issue-specific corpora \cite{card2015media, hamborg2019biascorp}. Such proxies fail to capture how individual readers perceive bias within the same article.

Subjective interpretation is central to this task. Prior work has shown that perceptions of media slant vary with reader background, political orientation, and article structure \cite{budak2015fair, roy2021framing}. Yet few datasets in NLP directly model this perception, especially at the document level, with full context and justification. Some efforts annotate sentence-level bias \cite{hamborg2019biascorp} or analyze framing in limited topics \cite{card2015media}, while others simulate reader opinion using weak supervision \cite{roy2021framing}. However, there remains a lack of high-quality, article-level resources that capture perceived bias, account for inter-annotator disagreement, and support explainability through rationale \cite{fairnessmedia, studentnewsdaily}.

We introduce \dataset, a dataset of 300 U.S. political news articles annotated for perceived ideological slant. Articles are sampled from a curated pool of 900 partisan news stories covering diverse political events. Each article is labeled by crowdworkers along two independent Likert scales: one for sentiment toward the Democratic Party, and another toward the Republican Party. In addition to these dual-axis labels, annotators select rationale indicators adapted from media literacy frameworks \cite{fairnessmedia, studentnewsdaily} and highlight relevant snippets. These rationales serve as built-in explanations, enabling downstream models not only to detect bias but also to justify their predictions. A qualification test and pilot-phase refinement ensure annotation quality \cite{Akkaya2010AmazonMT}. Table \ref{tab:annotation-example} shows a concrete example illustrating annotation schema and rationale tags.

\begin{table}[t]
\centering
\begin{tcolorbox}[title=Example Annotation Entry, colback=gray!5, colframe=black!40!black, boxrule=0.4pt]
\textbf{Title:} \textit{Anti-Trump celebs plan 'People's State of the Union' event} \\
\textbf{Event:} President Trump will deliver his first State of the Union \\[0.3em]

\textbf{Article Snippet (excerpt):} \\
\textit{A group of \textbf{Hollywood elites}, progressive groups, and other Trump opponents are planning a “People’s State of the Union” to counter the president’s first official address. The event, coordinated by unions, organizers of the Women’s March and Planned Parenthood, is being marketed as a celebration of the “resistance,” closer to “the people’s point of view,” USA Today reported.} \\[0.5em]

\textbf{Marked Bias Indicators:} 
\begin{itemize}
    \item \textbf{Marginalization of one side} (Indicator 4): \textit{“A group of Hollywood elites … celebration of the resistance”}
    \item \textbf{Emotionally charged language} (Indicator 0): \textit{“Hollywood elites,” “social activists,” “public alternative”}
\end{itemize}

\textbf{Worker Labels:} Right, Right \\
\textbf{Final Human Label:} \textbf{Right} \\
\textbf{Outlet Bias:} Right
\end{tcolorbox}
\caption{An example article from the dataset showing final labels, annotator agreement, and highlighted rationale types.}
\label{tab:annotation-example}
\end{table}

To benchmark the task beyond human annotation, we simulate labeling using GPT-4o under the same schema. 
This enables direct comparison of human and model judgments, revealing consistent asymmetries in perceived bias, most notably, a shared tendency to interpret subtly right-leaning content as neutral or even left-leaning. 
While recent work probes LLM–human alignment across moral reasoning \cite{Garcia2024MoralTuring}, preference-guided evaluation \cite{Liu2024PairwisePref}, and grammaticality judgments \cite{Hu2024GrammarAlign}, BiasLab complements these by modeling alignment and misalignment of LLMs and human feedback in the domain of political news.

We also define a \emph{perception drift} task, measuring alignment between perceived bias and outlet ideology, and establish a baseline for \textit{rationale prediction}, highlighting the need for richer contextual or multi-view representations.

BiasLab offers more than raw annotations: its structured schema transforms political bias perception into a testbed for human–model alignment. By capturing both directional sentiment and annotated rationale, it enables models to learn not just what bias humans perceive, but why. This positions BiasLab as a feedback-rich benchmark for training and evaluating alignment methods that go beyond label replication, grounded in public perception rather than rigid source-level ideology.

\section{Related Work}

Detecting political bias in news media has attracted considerable interest across computational linguistics, journalism studies, and political science. Existing research spans manual assessments at the source level to detailed analyses, including sentence-level annotations and event-specific framing studies. Prominent manual rating systems, such as the AllSides Media Bias Ratings \cite{allsides} and the Ad Fontes Media Bias Chart \cite{adfontes}, provide widely-used outlet-level ideological classifications but lack granularity at the article or sentence levels.

Computational methods have expanded bias detection efforts to more detailed annotations within news texts. The Media Frames Corpus \cite{card2015media} annotates framing techniques related to specific political issues but does not directly address ideological bias labeling. Hamborg et al.’s WCL corpus \cite{hamborg2019biascorp} offers sentence-level bias annotations; however, its scale and granularity restrict broader applicability at the document level.

Recently Machine learning approaches have significantly advanced political ideology detection. Iyyer et al. \cite{iyyer2014political} utilized recursive neural networks to predict political ideology based on congressional speeches, a distinct context from general news media. Roy et al. \cite{roy2021framing} investigated framing identification in immigration discussions on social media, but their approach did not explicitly provide supervised labels at the article level.

Crowdsourcing has emerged as an effective methodology for subjective annotation tasks, including bias detection. Budak et al. \cite{budak2015fair} demonstrated the effectiveness of crowd workers in quantifying media bias, emphasizing scalability and inter-annotator reliability. Nonetheless, challenges related to annotator agreement and managing subjective perceptions remain prevalent \cite{Akkaya2010AmazonMT, budak2015fair}. 
Recent work has begun to scrutinize LLM–human alignment across several dimensions. Garcia et al.\ introduce a Moral Turing Test, revealing systematic gaps between human moral choices and LLM responses \cite{Garcia2024MoralTuring}. Liu et al.\ demonstrate that incorporating pairwise human-preference feedback markedly improves an evaluator model’s agreement with human bias judgments \cite{Liu2024PairwisePref}. Complementing these findings, Hu et al.\ report strong model–human alignment on key grammatical constructions, suggesting that alignment quality can vary sharply by linguistic phenomenon \cite{Hu2024GrammarAlign}. Closer to our application domain, Baheti et al.\ analyze stance and safety interventions in offensive open-domain dialogue, highlighting practical challenges for bias-aware text generation \cite{Baheti2021toxiChat}. While these studies chart important territory in alignment, document-level political perception remains under-explored; our work fills this gap by benchmarking schema-constrained GPT-4o labels against human dual-axis bias annotations in real-world news.

While scalar preference signals, such as pairwise comparisons \cite{Liu2024PairwisePref} are widely used in alignment pipelines, recent work has explored more structured forms of human feedback, including multi‐choice critiques \cite{christiano2017deep} and preference rankings \cite{ouyang2022instruct}. \dataset ~complements these approaches by providing fine-grained rationales, label-reason mapping, flexible multi‐dimensional ratings, and explanatory feedback, and
can serve as interpretable supervision for alignment tasks in politically sensitive domains.


The \dataset ~dataset presented here extends the literature by explicitly targeting article-level bias annotations using dual-axis assessments of sentiment toward Democratic and Republican parties, thus addressing ambiguities prevalent in single-axis evaluations. The dataset employs structured rationale indicators, rigorous annotator qualification, and quality assurance processes, enhancing explainability and reliability. Unlike previous datasets, \dataset ~specifically concentrates on polarized sources, deliberately excluding centrist content to focus on the explicit complexities of ideological polarization. Furthermore, by directly comparing human annotations with schema-constrained language model-generated annotations, \dataset ~uniquely investigates human and automated perceptions of bias, documenting asymmetries and perception drift phenomena.

\section{Dataset Construction} 

\subsection{Article Collection and Source Bias Assignment}

\begin{figure}[ht]
    \centering
    \includegraphics[width=1\linewidth]{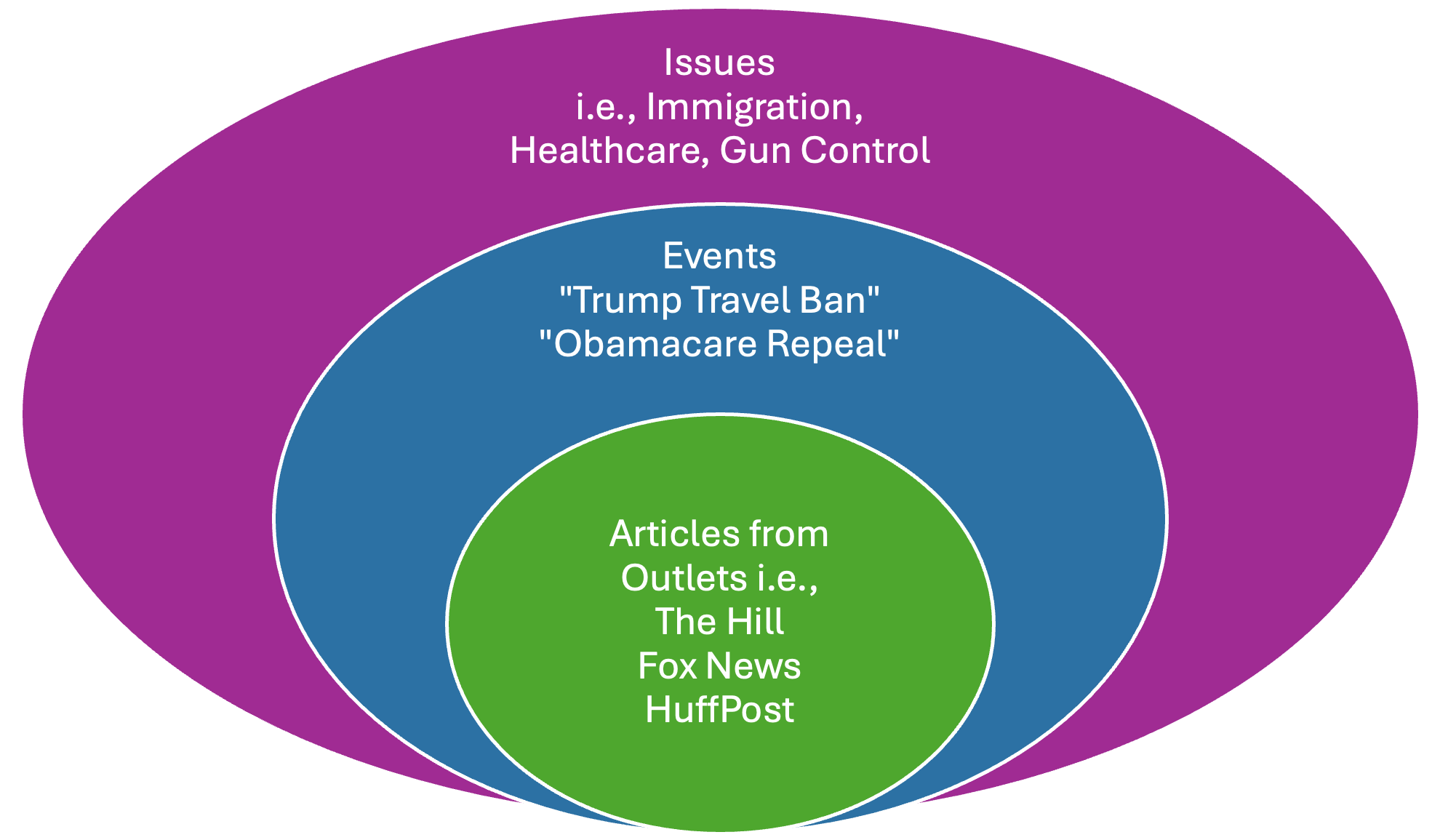}
    \caption{Conceptual hierarchy of data structure in \dataset.}
    \label{fig:data-structure}
\end{figure}

We collected over 10,000 political news articles from a wide range of online media sources spanning the ideological spectrum. 
Articles were retrieved via a structured web crawling and metadata extraction pipeline, 
%
Articles were aligned to more than 2000 real-world political events, which are grouped under broader issue categories. Figure \ref{fig:data-structure} shows the conceptual hierarchy of data structure in our dataset. Events were categorized into 50+ high-level issues (e.g., Healthcare, Immigration, FBI, Economy), reflecting areas of expected partisan divergence. Event distribution for top 20 issues is shown in Appendix, Figure~\ref{fig:event-distribution}. 


\subsection{Curated Subset and MTurk Task Preparation}

From this large collection, we manually curated a dataset of 900 political news articles drawn from 414 distinct events.
The selection criteria included balance of topic coverage, known standing of the outlets, and narrative clarity. 
Events included contentious legislative debates, political rallies, protests, and high-profile political developments between 2016–2018.
Each article was linked to its source event and assigned an outlet-level bias using third-party assessments from AllSides \cite{allsides} and Ad Fontes Media \cite{adfontes}, identifying each as Left, Lean Left, Lean Right, or Right. While the full dataset includes articles from centrist outlets, the 900 curated subset selected for annotation contains only clearly partisan sources (i.e., Left or Right aligned).
%
We then selected 300 articles from this curated pool for crowd-sourced annotation via Amazon Mechanical Turk (MTurk). This final set forms the core of the BiasLab dataset and is released alongside metadata for the broader 900-article pool to support future annotation efforts and replication studies.
Among the 300 articles, 202 came from outlets considered \textit{Right}-aligned or \textit{Leaning Right} and 98 of them came from outlets considered \textit{Left}-aligned or \textit{Leaning Left}.

\section{Annotation Protocol and Feedback Schema}
\subsection{Crowdsourced Annotation Task Design}
\begin{figure}[h]
 \centering
\includegraphics[width=0.48\textwidth]{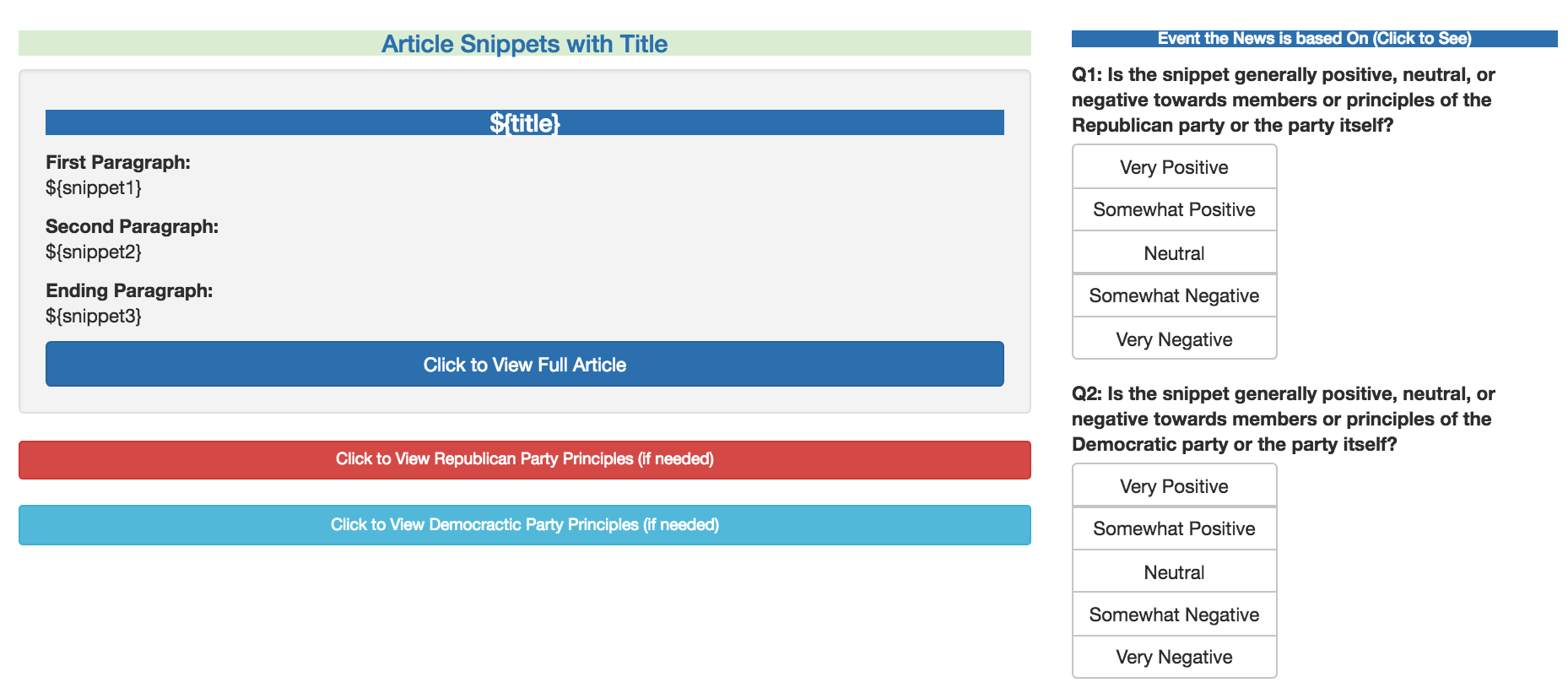}
\caption{Bias Identification Task}
\label{fig:mturkq}
\end{figure}

%
To capture nuanced ideological bias while minimizing annotator fatigue, we designed a structured annotation task on Amazon Mechanical Turk (MTurk). Annotators were shown each article as a sequence of three representative snippets: the title and lead paragraph, a middle paragraph, and the conclusion.
This design was informed by pilot findings showing that ideological framing is often concentrated near the beginning and end of political articles. Annotators also had access to the full article text via expandable toggle and article title, although snippets were the main annotation focus.
Figure \ref{fig:mturkq} shows the annotation interface deployed on MTurk. 



Rather than directly asking whether an article was `biased', we adopted a dual-axis perception framework.
Annotators independently rated the article’s tone toward:

\begin{enumerate}
    \item The Republican Party and its principles
    \item The Democratic Party and its principles
\end{enumerate}

Each question used a five-point Likert scale: \emph{very positive, somewhat positive, neutral, somewhat negative, very negative}.

To support interpretability and explainability, a third question asked annotators to explain their ratings by selecting from a list of predefined \textbf{bias indicators} adapted from media literacy frameworks \cite{fairnessmedia,studentnewsdaily}:
%
%

\begin{itemize}
    \item Strong emotionally charged language.
    \item Author opinion inserted as fact.
    \item Positive or negative presentation of political principles.
    \item Marginalization or omission of one side.
    \item Vague or one-sided sourcing.
    \item Inaccurate or incomplete evidence.
    \item The article was neutral and factual.
\end{itemize}

The seven rationale types align with three overarching dimensions of media bias: \textit{language and framing, balance and fairness,} and \textit{factual integrity}, as summarized in Appendix Figure~\ref{fig:bias-taxonomy}.
Annotators could select multiple reasons and were instructed to \textbf{highlight relevant text in the snippets} to justify their selections. Annotators could optionally provide a brief rationale for their selections.
Table \ref{tab:annotation-example} presents a representative annotated article with marked bias indicators and final labels.
To support detection of `\textit{omission bias}', we provided a short, neutral event description and summaries of both parties' core stances.
%

\subsection{Annotator Qualification and Quality Assurance}

To ensure label quality, we restricted participation to U.S.-based workers with at least 1,000 approved HITs and an approval rating above 96\%. Before accessing the main task, workers were required to pass a qualification test designed to assess their ability to recognize different forms of ideological bias. Use of qualifications tests to remove spammers and low quality
workers is a common and effective practice on Mechanical Turk \cite{Akkaya2010AmazonMT}.

\begin{figure}[ht]
 \centering
\includegraphics[width=0.48\textwidth]{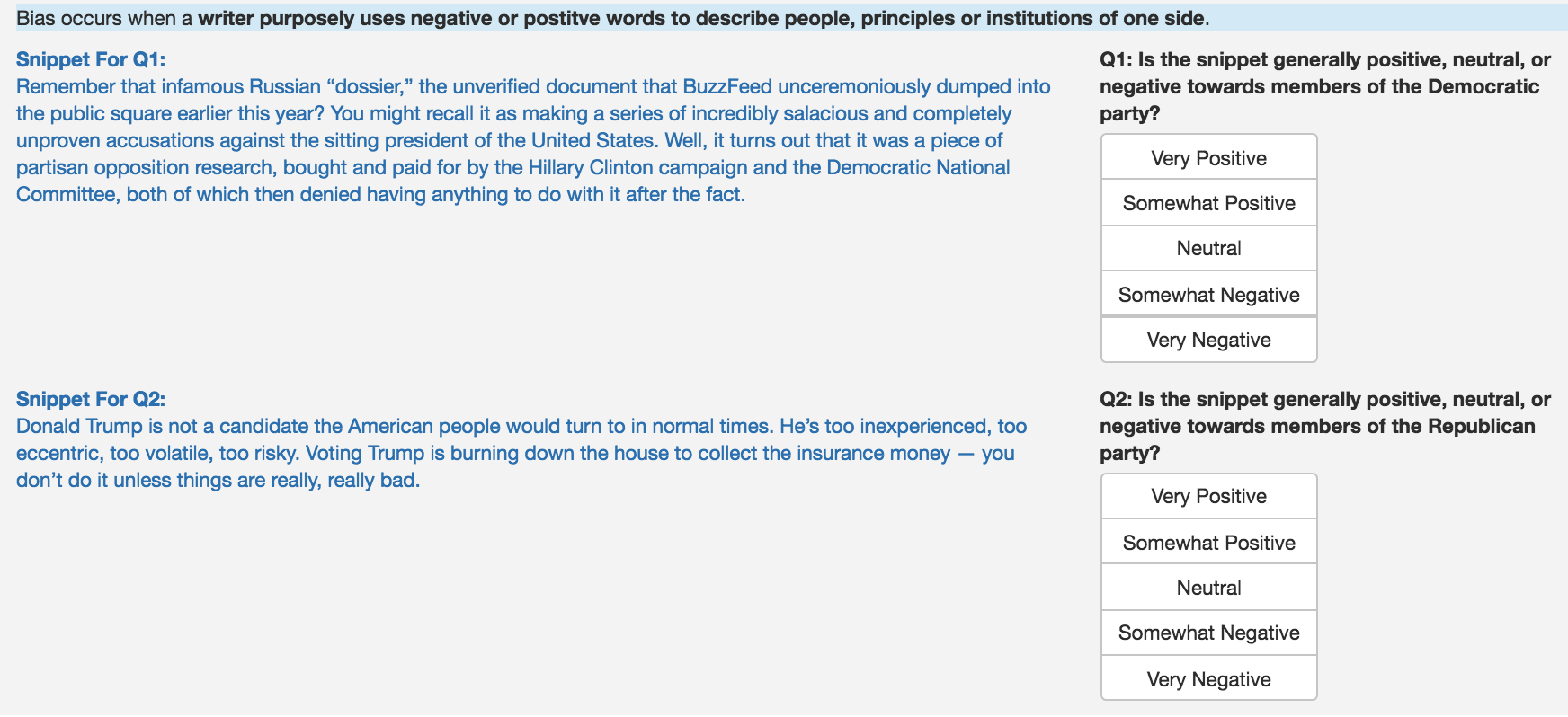}
\caption{Qualification Test Questions}
\label{fig:testq1}
\end{figure}

The test included seven multiple-choice questions that presented excerpts from news articles and asked workers to identify types of bias present. Each question included detailed feedback explaining the correct answer, serving both as a training step and a filter to ensure baseline annotation competency. As shown in figure \ref{fig:testq1}, we added two questions for testing the capability of identifying \textit{`Bias by Word Choice and Tone'}.

This qualification phase was informed by an earlier pilot round, where we observed substantial disagreement among annotators in identifying implicit forms of bias such as framing, omission, spin, or tone. 
The resulting test helped filter out inattentive or unqualified annotators and find workers with a baseline understanding of political media narratives.



\subsection{Lessons from Pilot Phase}
Before finalizing our annotation pipeline, we conducted a pilot phase involving 16 political news articles, each annotated by five independent MTurk workers. This phase was used to surface disagreement patterns and evaluate early task clarity. Two domain experts manually reviewed the resulting labels, particularly in cases of strong inter-annotator divergence, to identify the causes of confusion.
This analysis surfaced several core challenges:

\begin{itemize}
    \item \textbf{Ambiguity in inferred bias}: Annotators often disagreed when bias was not explicitly stated but implied through tone, framing, or omission.
    
    \item \textbf{Length-induced cognitive load}: Multi-paragraph passages led to fatigue and inconsistent labels.
    
    \item \textbf{Asymmetric political expectations}: 
    Some annotators over-identified bias based on topic alone (e.g., interpreting any Trump criticism as left-leaning).
\end{itemize}

The refinements described in earlier subsections
were directly informed by these pilot-phase findings.
We include the 16-pilot-document set and expert annotations in the release for transparency and methodological reproducibility.

\section{Feedback Quality and Human–Model Alignment}
To derive reliable gold labels from crowd annotations, we evaluated inter-annotator agreement and analyzed how perceived bias labels aligned with source-level ideological ratings. 
\textbf{Each of the 300 annotated articles was labeled by two independent MTurk workers}, and final labels were assigned based on a tiered resolution strategy. These labels were further compared to the expected bias of the publishing outlet to assess annotator tendencies and construct evaluation subsets for supervised tasks.

\subsection{Label Aggregation and Conflict Resolution Strategy}
For each news article, final human-perceived bias labels were assigned based on agreement between the two MTurkers using a three-way resolution scheme: 
\begin{itemize}
    \item Full agreement: agreed partisan label used.
    \item Partial agreement (Center + partisan): partisan label used.
    \item Full disagreement (Left vs. Right): marked as a conflict. 
\end{itemize}

This choice reflects two practical considerations: (1) center labels in such cases often reflect uncertainty rather than neutrality, and (2) at least one worker showed a pattern of overusing `Center', likely as a safe default. To preserve specificity and reduce noise, we prioritized the more directional label in these cases.
Our gold dataset also includes a final partisan score which is an average of the likert scale numbers from the two questions in the task. Negative values for the final partisan score indicated a left bias and positive values indicated a right bias for the article. Finally, a score with $0 \pm 0.5$ was considered centrist.

%

\subsection{Annotator-Outlet Agreement and Conflict Analysis}
%
To better understand how perceived bias aligns with the source outlet’s known political leaning, we conducted a post-hoc analysis of the annotations. 
We categorized the results into four groups, as summarized in Table \ref{tab:agreement_summary}:

\begin{table*}[ht]
\centering
\caption{Summary of human annotation agreement versus outlet bias}
\begin{tabular}{|l|c|l|}
\hline
\textbf{Category} & \textbf{Count} & \textbf{Possible Use Cases} \\
\hline
Workers agreed and \textbf{matched} outlet bias & 144 & Training (strong supervision) \\
Workers agreed on \textbf{Center}, outlet bias was Left/Right & 72 & Modeling perceptual neutrality \\
Workers agreed, but \textbf{mismatched} with outlet bias & 54 & Stress-testing source-level models \\
Workers \textbf{disagreed} with each other & 30 & Hold-out / qualitative analysis \\
\hline
Total & 300 &  \\
\hline
\end{tabular}
\label{tab:agreement_summary}
\end{table*}


\begin{itemize}
    \item A \textbf{majority (48\%)} of documents showed agreement between annotators and alignment with the outlet's known bias.
    \item In 72 cases, annotators agreed that the article was neutral or centrist, despite being from a partisan outlet. This suggests either subtle or balanced framing.
    \item In 54 cases, annotators agreed on a label that conflicted with the outlet’s bias, indicating either atypical reporting or perception drift.
    \item 30 documents had inter-annotator disagreement, underscoring inherent subjectivity in political bias perception.
\end{itemize}

This analysis supports our claim that \textbf{human-perceived bias can diverge from source-level assumptions}, and emphasizes the value of capturing both agreement and divergence in annotation pipelines.

\subsection{Quantitative Alignment with Outlet Bias}
\begin{figure}[!htbp]
    \centering
    \includegraphics[width=0.9\linewidth]{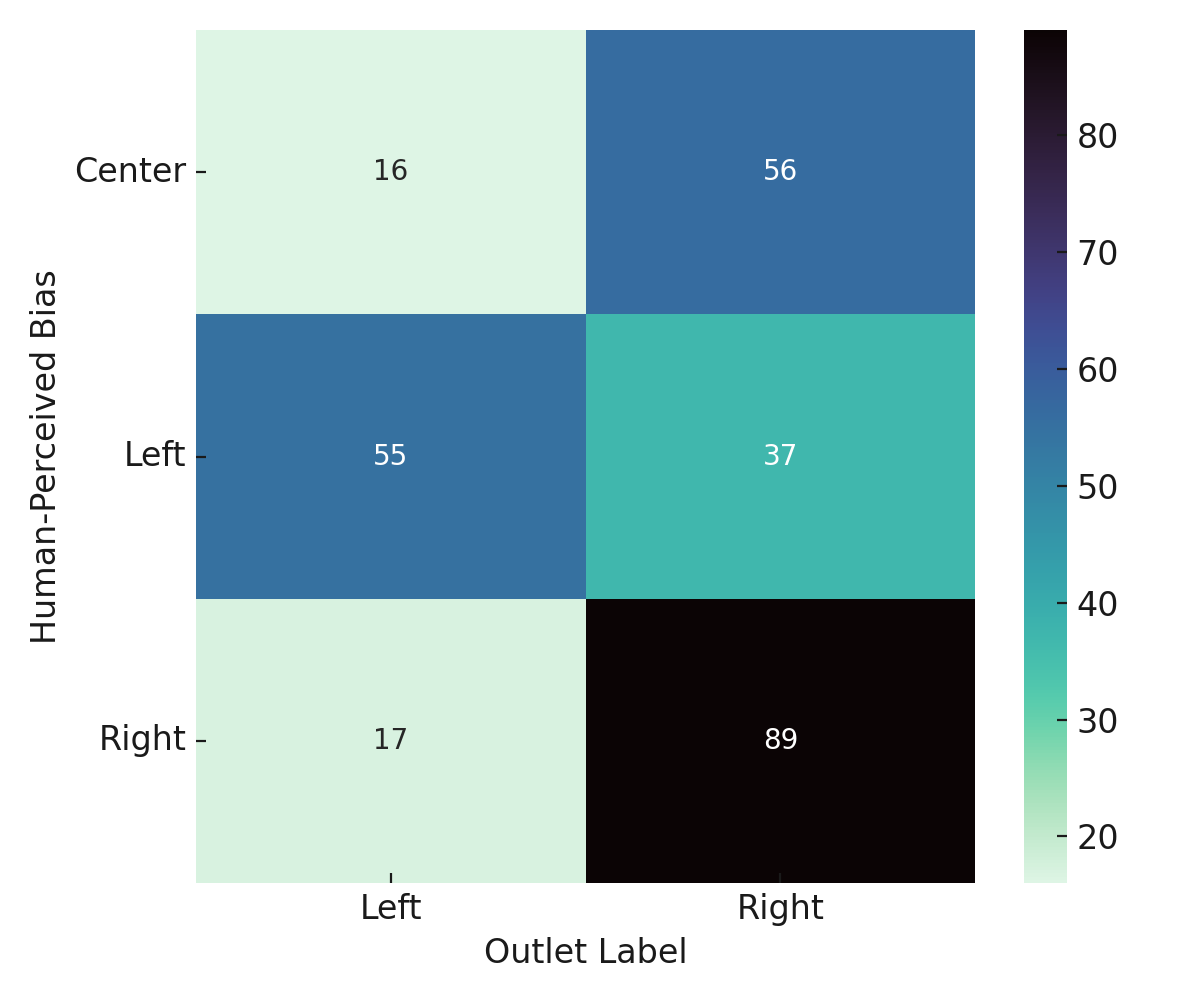}
    \caption{Confusion matrix of human-perceived bias vs. outlet bias}
    \label{fig:confmat}
\end{figure}

To evaluate the reliability of the annotation, we compared the final human-perceived bias labels for each article
to it's source ideology.  
While source-bias is an imperfect ground truth, it provides a
coarse measure of annotation accuracy under the assumption
that most articles reflect their outlet’s ideological leaning. The confusion matrix (Figure~\ref{fig:confmat}) illustrates this alignment across the 270 documents where MTurkers agreed. 

\begin{table}[ht]
\centering
\caption{Per-Class Performance (Human Label vs. Outlet Bias)}
\begin{tabular}{l|c|c|c}
\toprule
\textbf{Class} & \textbf{Precision} & \textbf{Recall} & \textbf{F1-score} \\
\midrule
Left     & 0.59 & 0.56 & 0.57 \\
Right    & 0.84 & 0.44 & 0.58 \\
Center   & 0.00 & 0.00 & 0.00 \\
\midrule
Macro Avg & 0.48 & 0.33 & 0.41 \\
Weighted Avg & 0.76 & 0.53 & 0.62 \\
\bottomrule
\end{tabular}
\label{tab:prf1}
\end{table}

 Table~\ref{tab:prf1} summarizes per-class performance metrics for the same 270 articles.
While source bias serves only as a weak supervisory signal, the class-level scores illustrate both alignment and divergence trends. Annotators performed most consistently on articles from right-leaning outlets (F1 = 0.58), likely due to more overt partisan cues in those articles. In contrast, left-leaning articles yielded more balanced precision and recall, while the Center class appears with zero scores. 

Notably, \textbf{this absence of Center-class predictions reflects our dataset design}, not annotator behavior: all 300 articles were drawn from sources categorized as Left, Lean Left, Right, or Lean Right, and \textbf{no articles from centrist outlets were included in this release}. As a result, the confusion matrix and performance metrics for Center offer no insight into labeling tendencies and should not be interpreted as annotator bias.



\subsection{Comparison with GPT-4o Annotations}
To further evaluate label consistency and annotator reliability, we conducted a controlled simulation using OpenAI’s GPT-4o model. The model was prompted with our \textit{BiasLabelQualification} schema and constrained to the same information seen by MTurk workers: only the article title and three snippets. No event label or full document was provided. In contrast to MTurk’s multiple annotators, we simulated a single OpenAI annotator agent, and used the scores from both left and right Likert scales to finalize the manual label.

GPT-4o was tasked with generating scores for all three original questions in the MTurk design as well as a final bias label for all 300 articles in our manually annotated set. These outputs were compared against the outlet-based bias labels.

\begin{itemize}
    \item \textbf{Agreement} with outlet label: 59\%     
    \item This exceeds the 48\% agreement achieved by MTurk workers matching outlet labels. 
    \item The model’s most common error was labeling Right articles as Left, which is also observed among human annotators, but humans lean towards Center more.
\end{itemize}

\begin{figure}[!htbp]
    \centering
    \includegraphics[width=0.9\linewidth]{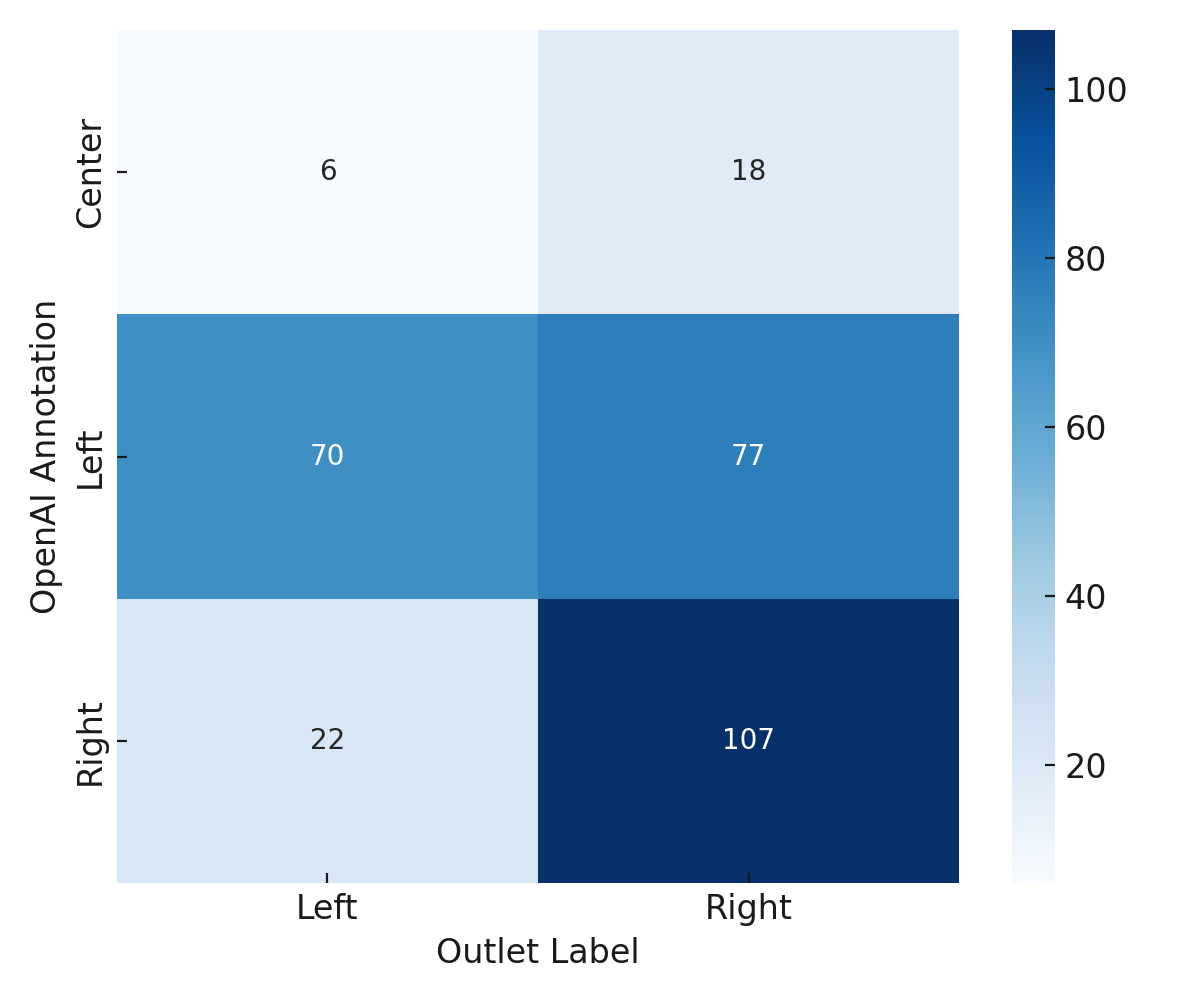}
    \caption{Confusion matrix of GPT-4o-generated bias labels vs. outlet-provided labels.}
    \label{fig:conf-matrix-openai}
\end{figure}
\begin{figure}[!htbp]
    \centering
    \includegraphics[width=0.9\linewidth]{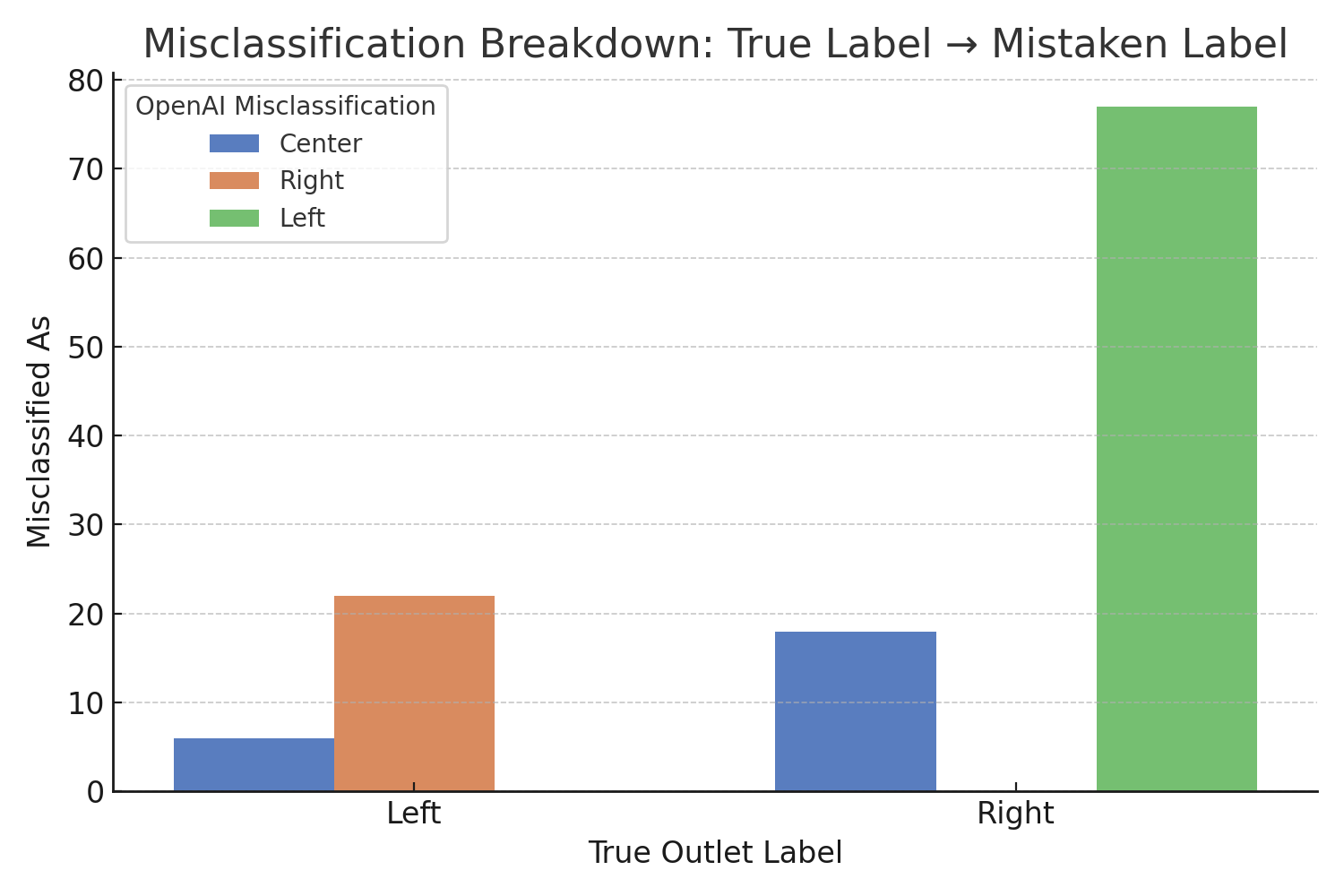}
    \caption{Error patterns showing most frequent misclassifications (e.g., Left → Right)}
    \label{fig:error-patterns-openai}
\end{figure}
%

Figure \ref{fig:conf-matrix-openai} and \ref{fig:error-patterns-openai} show the error confusion across GPT-4o labels and common misclassification patterns. Batch-wise annotation accuracy for GPT-4o is included in the Appendix.  While the model outperformed average human agreement, it still struggled with subtle right-leaning framings and Center-vs-Right distinctions. 
These results reinforce both the difficulty and subjectivity of the bias labeling task, even for high-performing Large Language Models (LLMs).

\subsection{Final Gold Set}
We define a \textbf{gold-standard subset} of 270 articles derived from MTurk annotations, based on strict agreement criteria between independent annotators. This set is used for supervised learning, evaluation, and perception analysis:

\begin{itemize}
    \item \textbf{Agreed and matched outlet bias (144 docs)}: high-confidence supervision
    \item \textbf{Agreed as Center on partisan outlet (72 docs)}: perception-driven neutrality
    \item \textbf{Agreed but conflicted with outlet (54 docs)}: included as gold for contrastive evaluation and perception drift identification.
\end{itemize}

The remaining 30 conflict-labeled documents are excluded from gold set but retained as examples of annotation difficulty.

\paragraph{Auxiliary Labels.}
In addition to the human-annotated gold set, we release two supplementary label sources:
\begin{itemize}
    \item \textbf{Outlet-derived bias labels} for all 300 annotated articles and the broader \textbf{900-document corpus}, based on the political leanings of their news sources. These serve as weak supervision signals and enable source-aware modeling.
    \item Schema-constrained \textbf{GPT-4o annotations} for all 300 articles, designed to replicate the MTurk labeling scheme. While not part of the human gold set, these enable LLM–human agreement analysis and consistency benchmarking.
\end{itemize}

Together, these provide a flexible foundation for both perception-based learning and benchmark modeling across human, source, and machine perspectives.








\section{Feedback-Based Benchmark Models}

To evaluate the utility of BiasLab for downstream tasks, we explore two classification settings using the annotated subset of 300 documents. These pilot experiments use simple baselines to demonstrate the feasibility and challenge of modeling human-perceived political bias.

\subsection{Perception Drift Identification}
We study the problem of predicting agreement between human-perceived bias and the ideological leaning of the source outlet. This captures what we refer to as \textit{perception drift}-cases where annotators consistently perceived bias in a direction that differs from the outlet’s established political alignment.


We frame this as a binary classification task using 198 articles where annotators agreed on a partisan label. Of these, 144 matched the outlet’s bias (agreement), while 54 conflicted (drift). We train a Logistic Regression classifier using TF-IDF features extracted from article text.


As shown in Table~\ref{tab:match_classification_results}, the model achieved an overall accuracy of 55.6\%, with notably higher recall for match cases than drift cases.
The class imbalance and linguistic subtlety of drift examples likely contribute to the lower performance on mismatches. These results validate the difficulty of detecting perceived bias drift from surface lexical features alone and establish a benchmark for future work.

\begin{table}[h]
\centering
\caption{Logistic Regression performance on perception drift identification (TF-IDF)}
\begin{tabular}{lccc}
\toprule
Class & Precision & Recall & F1 \\
\midrule
Aligned (1) & 0.56 & 0.76 & 0.65 \\
Drift (0) & 0.53 & 0.32 & 0.40 \\
\bottomrule
\end{tabular}
\label{tab:match_classification_results}
\end{table}


\subsection{Bias Rationale Type Classification}

We also evaluate whether models can predict the types of rationale selected by annotators to justify their perceived bias. These rationale types align with established media literacy categories, e.g., directional framing, structural imbalance, or neutral presentation, and were selected from the original annotation task. Each document may be associated with multiple rationale types, making this a multi-label classification task.

We train one-vs-rest Logistic Regression models for each rationale type using TF-IDF features. Table~\ref{tab:rationale_logreg} summarizes macro-averaged results for the most common rationale types. The full per-class breakdown is shown in Appendix Table~\ref{tab:appendix_rationale_full}.

\begin{table}[h]
\centering
\caption{Macro-averaged Logistic Regression performance for rationale type classification.}
\begin{tabular}{l|ccc}
\toprule
\textbf{Rationale Type} & \textbf{Precision} & \textbf{Recall} & \textbf{F1 Score} \\
\midrule
neutral\_other  & 0.70 & 0.64 & 0.61 \\
structural     & 0.61 & 0.56 & 0.54 \\
directional    & 0.62 & 0.54 & 0.51 \\
\bottomrule
\end{tabular}
\label{tab:rationale_logreg}
\end{table}

These results suggest that some rationale types, particularly \textit{neutral\_other} and \textit{structural}, can be learned to a modest degree using surface lexical features alone. However, the relatively low recall and F1 scores for even the most common rationale types indicate the limitations of simple models, and reinforce the interpretive and context-dependent nature of bias perception. We plan to extend this task using contextual embeddings and weak supervision from the remaining 600 curated articles and the 10K outlet-labeled corpus.


\section{Discussion and Future Work}

\dataset ~provides a transparent, perception-focused resource for analyzing ideological bias in political news. Unlike prior datasets centered on outlet ratings or issue framing, \dataset ~offers article-level annotations grounded in reader sentiment toward both major U.S. political parties.

First, we observe that annotator agreement is strongest when bias cues are overt, and diverges when bias is conveyed through framing, omission, or emotional tone. This subjectivity reflects the nature of the task, and not a flaw. Capturing disagreement is essential for modeling real-world perception. BiasLab's dual-axis design and rationale indicators help structure this ambiguity, enabling both interpretability and fine-grained evaluation.

Second, our \textit{tiered label resolution strategy}, which includes matched, mismatched, and centrist-perceived labels, offers structured training subsets and supports contrastive studies. These distinctions are critical for evaluating not just bias detection accuracy but perception drift and labeling asymmetry.

Third, while BiasLab includes \textit{300 fully annotated articles}, it is curated from a 900-document partisan subset and a broader 10K article collection. This smaller release was prioritized for clarity and replicability, but our infrastructure supports easy expansion using the same MTurk protocol.

Additionally, we evaluated LLM-based labeling using GPT-4o under schema-constrained prompts. GPT-4o slightly outperformed human annotators on outlet agreement but mirrored human tendencies to misclassify subtle right-leaning content. 
These results reinforce that even high-performing LLMs struggle with nuance-sensitive, perception-driven tasks, and do not resolve the core ambiguity around ideological slant. 
This also highlights the limits of using output agreement as a proxy for interpretive alignment and points to the need for richer modeling approaches that account for ambiguity and rationale diversity.

We include initial modeling experiments to benchmark dataset utility and illustrate the difficulty of perception-driven classification. These pilots use interpretable, replicable classifiers, and we leave deeper model tuning for future work. Together, \dataset ~offers a replicable, perception-grounded platform for evaluating media bias, interpretability, and alignment between human and machine.

\section{Conclusion}
\dataset ~introduces a new resource for explainable political bias detection, pairing dual-axis sentiment labels with rationale indicators that surface why an article reads as left- or right-leaning. Our human-vs-LLM comparisons expose where both groups falter, especially on subtle framings, and our two baseline tasks (perception drift and rationale classification) underline the real-world difficulty of building transparent bias models. Moving forward, we plan to integrate richer contextual signals and interactive, human-in-the-loop workflows to strengthen explanation quality. 
We invite the community to explore new modeling approaches that treat structured human judgments as actionable alignment signals for perception-driven language model behavior.

\section{Limitations}
BiasLab has three key limitations. \textit{First}, centrist articles were excluded to focus on clear polarization. \textit{Second}, annotators reviewed snippets rather than full articles to reduce fatigue and increase annotation quality. \textit{Third}, we did not collect annotator demographics or political orientation, limiting fairness or bias assessments.



\section*{Acknowledgment}
This research builds upon an initial annotation methodology developed under the supervision of Dr.\ Dan Goldwasser during the author’s time at Purdue University. Dr.\ Goldwasser provided guidance and funding for the original MTurk pilot phase. The data collection, cleaning, schema formalization, full dataset release, LLM simulations, and all new analyses presented here were conducted solely by the author. 


\section*{Ethics Statement}
This research used publicly available political news articles and involved crowd-sourced annotations via Amazon Mechanical Turk. Workers were restricted to U.S.-based accounts with a strong approval history. A qualification test was used to filter for quality. No personal or sensitive data was collected.

\bibliographystyle{icml2025}

\begin{thebibliography}{00}
\bibitem{entman2007framing}
Robert M. Entman.  
\newblock Framing bias: Media in the distribution of power.  
\newblock \emph{Journal of Communication}, 57(1):163–173, 2007.

\bibitem{lazer2018fakenews}
David M. Lazer, Matthew Baum, Yochai Benkler, Adam J. Berinsky, Kelly M. Greenhill, Filippo Menczer, Miriam J. Metzger, Brendan Nyhan, Gordon Pennycook, David Rothschild, Michael Schudson, Steven A. Sloman, Cass R. Sunstein, Emily A. Thorson, Duncan J. Watts, and Jonathan L. Zittrain.  
\newblock The science of fake news.  
\newblock \emph{Science}, 359(6380):1094–1096, 2018.

\bibitem{allsides}
AllSides. Media Bias Ratings. \url{https://www.allsides.com/media-bias/media-bias-ratings}

\bibitem{adfontes}
Ad Fontes Media. Media Bias Chart. \url{https://adfontesmedia.com}

\bibitem{card2015media}
Dallas Card, Amber E. Boydstun, Justin H. Gross, Philip Resnik, and Noah A. Smith.  
\newblock The Media Frames Corpus: Annotations of Frames Across News Articles.  
\newblock In \emph{Proceedings of NAACL-HLT}, 2015.

\bibitem{hamborg2019biascorp}  
Felix Hamborg, Anastasia Zhukova, and Bela Gipp. 
\newblock Automated Identification of Media Bias by Word Choice and Labeling in News Articles.
\newblock In Proceedings of the 19th ACM/IEEE-CS Joint Conference on Digital Libraries (JCDL ’19), pages 196–205.


\bibitem{iyyer2014political}
Mohit Iyyer, Peter Enns, Jordan Boyd-Graber, and Philip Resnik.  
\newblock Political Ideology Detection Using Recursive Neural Networks.  
\newblock In \emph{Proceedings of ACL}, 2014.

\bibitem{roy2021framing}
Shamik Roy, Chenghao Liu, and Dan Goldwasser.  
\newblock Modeling Framing in Immigration Discourse on Social Media.  
\newblock In \emph{Proceedings of ACL}, 2021.

\bibitem{tsur2015framedetection}
Oren Tsur, Dan Calacci, and David Lazer,
\newblock “A Frame of Mind: Using Statistical Models for Detection of Framing and Agenda Setting Campaigns,”
\newblock in \emph{Proceedings of the 53rd Annual Meeting of the Association for Computational Linguistics and the 7th International Joint Conference on Natural Language Processing (Volume 1: Long Papers)}, July 2015, pp. 1620–1630.  

\bibitem{Akkaya2010AmazonMT}
C.~Akkaya, A.~Conrad, J.~Wiebe, and R.~Mihalcea.
\newblock Amazon Mechanical Turk for Subjectivity Word Sense Disambiguation.
\newblock In \emph{Proceedings of the NAACL HLT Workshop on Creating Speech and Language Data with Amazon's Mechanical Turk (Mturk@HLT-NAACL)}, 2010.

\bibitem{budak2015fair}
C.~Budak, S.~Goel, and J.~M. Rao.
\newblock Fair and Balanced? Quantifying Media Bias through Crowdsourced Content Analysis.
\newblock In \emph{Proceedings of the 2015 ACM Conference on Computer Supported Cooperative Work and Social Computing (CSCW)}, 2015.

\bibitem{fairnessmedia}
FAIR. How to detect bias in news media. 
\newblock \url{https://fair.org/take-action-now/media-activism-kit/how-to-detect-bias-in-news-media/}, accessed May 2025.

\bibitem{studentnewsdaily}
Student News Daily. Types of media bias.
\newblock \url{https://www.studentnewsdaily.com/types-of-media-bias/}, accessed May 2025.

\bibitem{Garcia2024MoralTuring}
Basile Garcia, Crystal Qian, and Stefano Palminteri.
\newblock The Moral Turing Test: Evaluating Human–LLM Alignment in Moral Decision-Making.
\newblock \url{https://arxiv.org/abs/2410.07304v1}

\bibitem{Liu2024PairwisePref}
Yinhong Liu, Han Zhou, Zhijiang Guo, Ehsan Shareghi, Ivan Vulić, Anna Korhonen, and Nigel Collier.
\newblock Aligning with Human Judgment: The Role of Pairwise Preference in Large Language Model Evaluators.
\newblock \url{https://arxiv.org/abs/2403.16950}

\bibitem{Hu2024GrammarAlign}
Jennifer Hu, Kyle Mahowald, Gary Lupyan, and Roger Levy.
\newblock Language Models Align with Human Judgments on Key Grammatical Constructions.
\newblock \emph{Proceedings of the National Academy of Sciences}, 121(21):e2400917121, 2024.
\newblock DOI:10.1073/pnas.2400917121
\newblock \url{https://www.pnas.org/doi/10.1073/pnas.2400917121}


\bibitem{Baheti2021toxiChat}
Ashutosh Baheti, Maarten Sap, Alan Ritter, and Mark Riedl.
\newblock Just Say No: Analyzing the Stance of Neural Dialogue Generation in Offensive Contexts.
\newblock \url{https://arxiv.org/abs/2108.11830}


\bibitem{ouyang2022instruct}
Long Ouyang, Jeffrey Wu, Xu Jiang, Diogo Almeida, Carroll Wainwright, Pamela Mishkin, Chong Zhang, Sandhini Agarwal, Katarina Slama, Alex Ray, John Schulman, Jacob Hilton, Fraser Kelton, Luke Miller, Maddie Simens, Amanda Askell, Peter Welinder, Paul Christiano, Jan Leike, and Ryan Lowe.
\newblock Training language models to follow instructions with human feedback.
\newblock In \emph{Advances in Neural Information Processing Systems}, volume 35, pages 27730--27744. Curran Associates, Inc., 2022.

\bibitem{christiano2017deep}
Paul Christiano, Jan Leike, Tom Brown, Miljan Martic, Shane Legg, and Dario Amodei.
\newblock Deep reinforcement learning from human preferences.
\newblock In \emph{Advances in Neural Information Processing Systems}, volume 30. Curran Associates, Inc., 2017.
\newblock \url{https://proceedings.neurips.cc/paper_files/paper/2017/file/d5e2c0adad503c91f91df240d0cd4e49-Paper.pdf}



\end{thebibliography}

\newpage
\appendix
\onecolumn

\section*{Appendix}

\begin{figure}[htbp]
    \centering
    \includegraphics[width=0.7\linewidth]{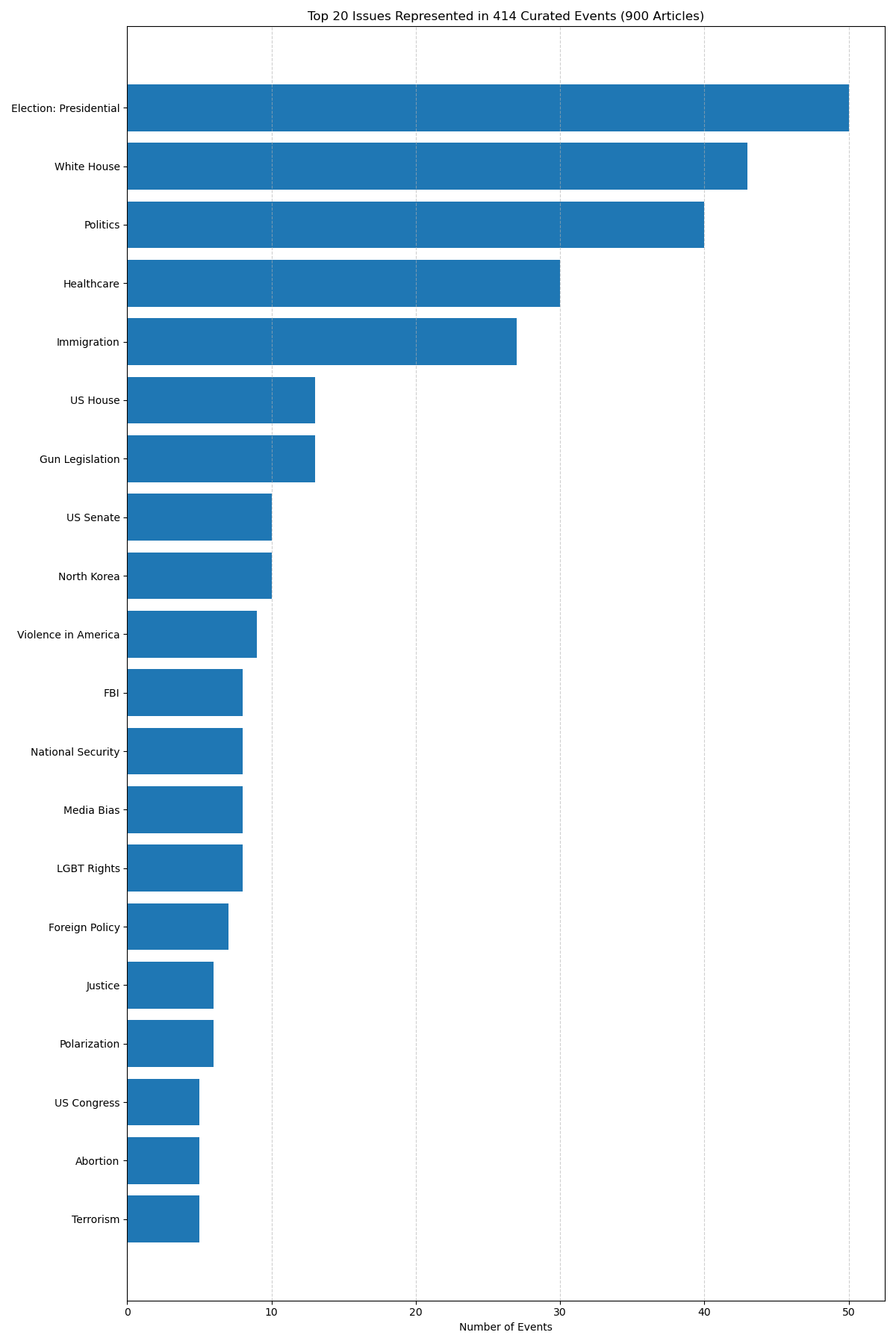}
    \caption{Event distribution across issues for the 414 events covering the 900 articles}
    \label{fig:event-distribution}
\end{figure}


\begin{figure}[htbp]
    \centering
    \includegraphics[width=\linewidth]{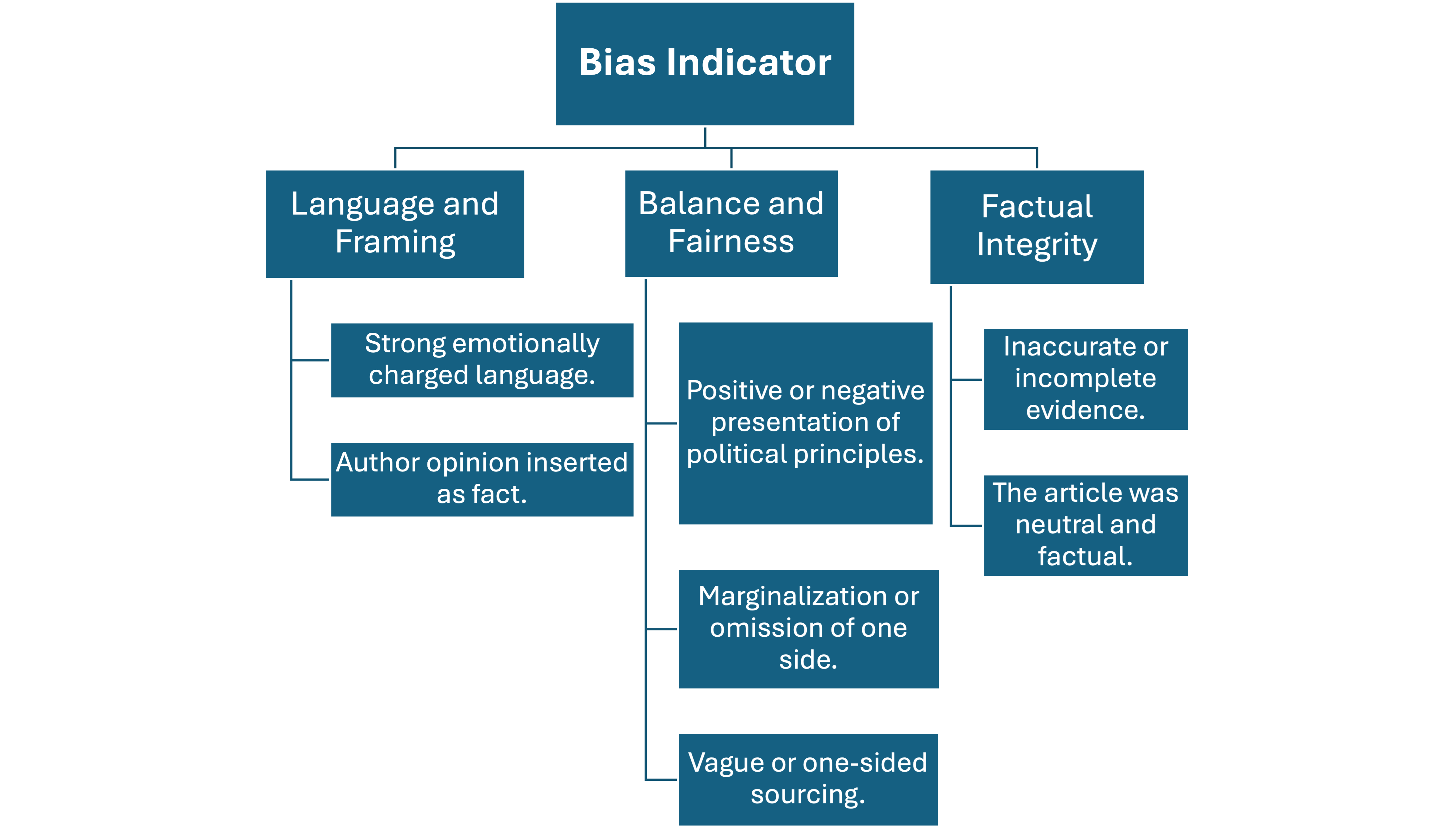}
    \caption{Hierarchical organization of bias rationale indicators used in annotation. The seven individual types are grouped into broader conceptual themes to support interpretation and modeling.}
    \label{fig:bias-taxonomy}
\end{figure}

\begin{figure}[htbp]
    \centering
    \includegraphics[width=1\linewidth]{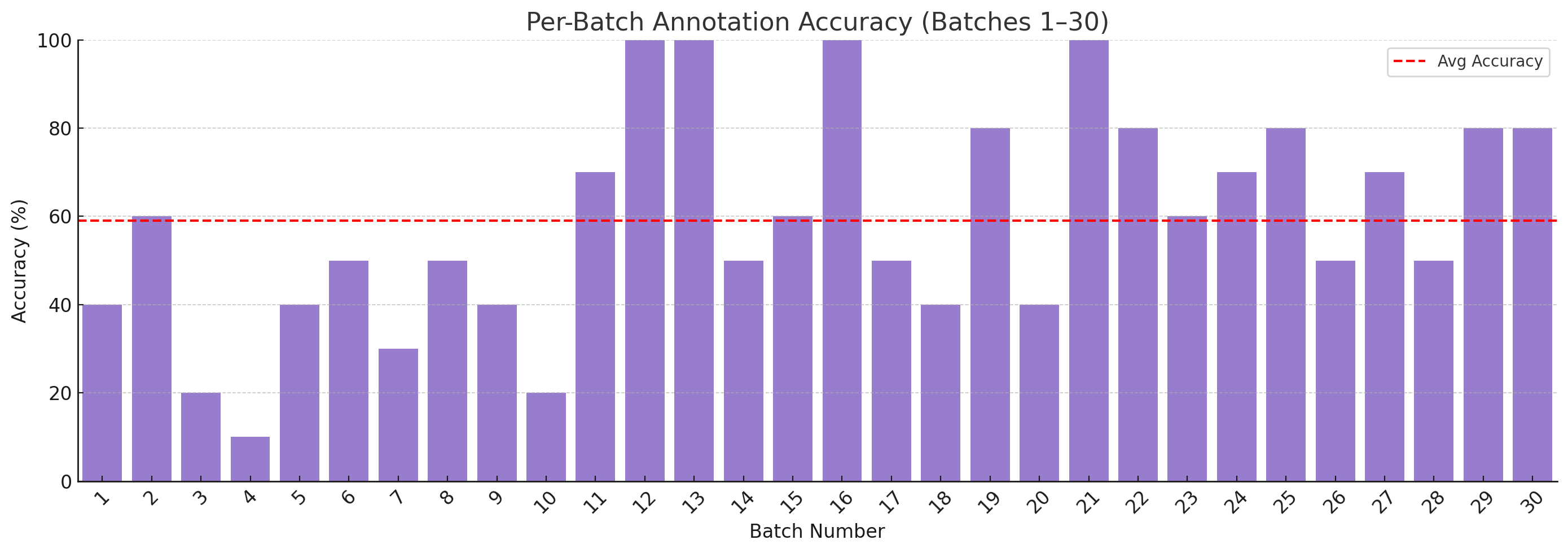}
    \caption{Batch-wise accuracy (\%) of GPT-4o annotations across 30 batches (300 articles).}
    \label{fig:batch-wise-accuracy-openai}
\end{figure}
\begin{table*}[htbp]
\centering
\caption{Full per-class Logistic Regression results for rationale type classification. Class 1 represents positive class (presence of rationale); Class 0 is the negative class.}
\begin{tabular}{llccc}
\toprule
\textbf{Rationale Type} & \textbf{Class} & \textbf{Precision} & \textbf{Recall} & \textbf{F1 Score} \\
\midrule
directional & class\_0 & 0.69 & 0.94 & 0.79 \\
directional & class\_1 & 0.55 & 0.15 & 0.24 \\
structural & class\_0 & 0.67 & 0.89 & 0.76 \\
structural & class\_1 & 0.56 & 0.23 & 0.32 \\
neutral\_other & class\_0 & 0.58 & 0.92 & 0.71 \\
neutral\_other & class\_1 & 0.81 & 0.36 & 0.5 \\
\bottomrule
\end{tabular}
\label{tab:appendix_rationale_full}
\end{table*}

\end{document}